\documentclass[sigconf, screen]{acmart}

\AtBeginDocument{%
  \providecommand\BibTeX{{%
    \normalfont B\kern-0.5em{\scshape i\kern-0.25em b}\kern-0.8em\TeX}}}

\copyrightyear{2023}
\acmYear{2023}
\setcopyright{rightsretained}
\acmConference[CUI '23]{ACM conference on Conversational User Interfaces}{July 19--21, 2023}{Eindhoven, Netherlands}
\acmBooktitle{ACM conference on Conversational User Interfaces (CUI '23), July 19--21, 2023, Eindhoven, Netherlands}
\acmDOI{10.1145/3571884.3604316}
\acmISBN{979-8-4007-0014-9/23/07}

\begin{document}

\title{Opening up ChatGPT: Tracking openness, transparency, and accountability in instruction-tuned text generators}



\author{Andreas Liesenfeld}
\email{andreas.liesenfeld@ru.nl}
\orcid{0000-0001-6076-4406}
\affiliation{
  \institution{Centre for Language Studies}
  \city{Radboud University}
  \country{The Netherlands}
 }
 
\author{Alianda Lopez}
\email{ada.lopez@ru.nl}
\orcid{0009-0004-5873-5496}
\affiliation{
  \institution{Centre for Language Studies}
  \city{Radboud University}
  \country{The Netherlands}
 }
\author{Mark Dingemanse}
\email{mark.dingemanse@ru.nl}
\orcid{0000-0002-3290-5723}
\affiliation{
  \institution{Centre for Language Studies}
  \city{Radboud University}
  \country{The Netherlands}
 }

\renewcommand{\shortauthors}{Liesenfeld, Lopez, and Dingemanse}

\begin{abstract}
Large language models that exhibit instruction-following behaviour represent one of the biggest recent upheavals in conversational interfaces, a trend in large part fuelled by the release of OpenAI's ChatGPT, a proprietary large language model for text generation fine-tuned through reinforcement learning from human feedback (LLM+RLHF). We review the risks of relying on proprietary software and survey the first crop of open-source projects of comparable architecture and functionality. The main contribution of this paper is to show that openness is differentiated, and to offer scientific documentation of degrees of openness in this fast-moving field. We evaluate projects in terms of openness of code, training data, model weights, RLHF data, licensing, scientific documentation, and access methods. We find that while there is a fast-growing list of projects billing themselves as `open source', many inherit undocumented data of dubious legality, few share the all-important instruction-tuning (a key site where human annotation labour is involved), and careful scientific documentation is exceedingly rare. Degrees of openness are relevant to fairness and accountability at all points, from data collection and curation to model architecture, and from training and fine-tuning to release and deployment. 


\end{abstract}

\ccsdesc[500]{Natural language generation}
\ccsdesc[300]{Emerging technologies}
\ccsdesc[300]{Surveys and overview}
\ccsdesc[100]{Open-source software}
\ccsdesc[100]{Evaluation}

\keywords{open source, survey, chatGPT, large language models, RLHF}


\begin{teaserfigure}
  \includegraphics[width=\textwidth]{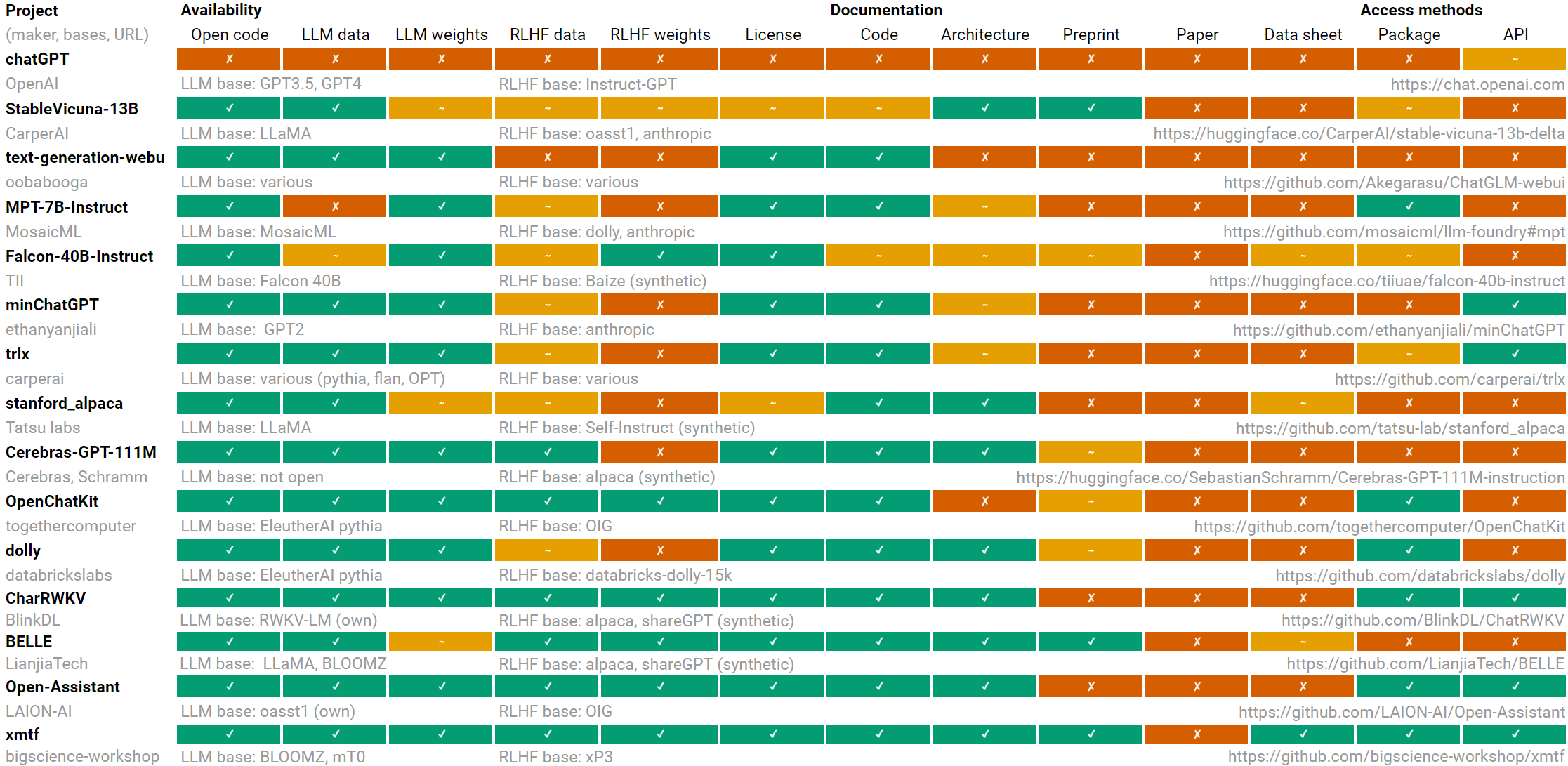}
  \caption{\textmd{How open and transparent are current instruction-following text generators? Snapshot as of June 2023 — updates at \href{http://osf.io/d6fsr}{osf.io/d6fsr}}}
  \Description{A table with 16 rows and 13 columns. The first row is headed ``Project" and lists the project names and organization behind it. Some projects also feature more information regarding the base large language and reinforcement learning models that are used. The The remaining 12 rows are each names after one of evaluation features in Table 1. Each cell of the table then evaluates the project for the respective feature, either giving it a pass, a partial pass, or a fail. More detailed information as well as the content of each cell can be found in the data repository that accompanies the paper.}
  \label{fig:teaser}
\end{teaserfigure}


\received{20 April 2023}
\received[accepted]{26 May 2023}


\maketitle

\section{Introduction}

Open research is the lifeblood of cumulative progress in science and engineering. In today's technological landscape, it is hard to find any research finding or technology that does not rely to a significant extent on the fruits of open research, often publicly funded. For instance, AlexNet \cite{krizhevsky_imagenet_2017}, the deep neural net kickstarting the deep learning revolution a decade ago, derived its strength from a human-annotated dataset of 3.2 million images created by Princeton computer scientists \cite{deng_imagenet_2009,chawla_ten_2023}. And the striking progress in protein folding in recent years (with the AlphaFold deep learning system predicting the structure of nearly all known proteins \cite{tunyasuvunakool_highly_2021}, where decades of prior work had reached a comparatively meagre 17\%) has only been possible thanks to openly deposited structural data in the Protein Data Bank that goes back half a century  \cite{bernstein_protein_1977}. 

The talk of the town in conversational interfaces today is undoubtedly ChatGPT, an instruction-tuned text generator that impresses many because of its fluid prose. Yet striking new capabilities should not detract us from the risks of proprietary systems. Only three months after OpenAI rolled out ChatGPT, it abruptly discontinued API support for its widely used Codex model that had been available as a ``free limited beta'' since 2021 \cite{pandey_openai_2023}  --- surprising users with only three days' notice and undercutting at one blow the reproducibility of at least 100 research papers.\footnote{See \href{https://aclanthology.org/search/?q=openai-davinci-002}{aclanthology.org/search/?q=openai-davinci-002} (the same search term yields >150 arXiv preprints and >800 entries on Google Scholar) } This is a stark reminder that proprietary systems are designed to offer smooth onboarding and convenience but come at the price of user lock-in and a lack of reliability. 

Proprietary systems come with considerable further risks and harms \cite{bender_dangers_2021,carlini_extracting_2021}. They tend to be developed without transparent ethical oversight, and are typically rolled out with profit motives that incentivise generating hype over enabling careful scientific work. They allow companies to mask exploitative labour practices, privacy implications \cite{larson_not_2021} and murky copyright situations \cite{schaul_inside_2023}. Today there is a growing division between global academia and the handful of firms who wield the computational resources required for training large language models. This ``Compute Divide'' \cite{ahmed_growing_2023} contributes to the growing de-democratisation of AI. Against this, working scientists call for avoiding the lure of proprietary models \cite{spirling_why_2023}, for decolonizing the computational sciences \cite{birhane_towards_2021}, and for regulatory efforts to counteract harmful impacts \cite{gebru_statement_2023}.

\subsection{Why openness matters}
Open data is only one aspect of open research; open code, open models, open documentation, and open licenses are other crucial elements \cite{gundersen_reproducible_2018,burgelman_open_2019}. Openness promotes transparency, reproducibility, and quality control; all features that are prequisites for supporting robust scientific inference \cite{mckiernan_how_2016}  and building trustworthy AI \cite{li_trustworthy_2023}. Openness also allows critical use in research and teaching. For instance, it enables the painstaking labour of documenting ethical problems in existing datasets \cite{schaul_inside_2023,birhane_multimodal_2021}, important work that can sometimes result in the retraction of such datasets \cite{birhane_large_2021}. In teaching, it can help foster critical computational literacy \cite{lee_none_2016}. 

Despite strong evidence of the scientific and engineering benefits of open research practices, openness is not a given in machine learning and AI research \cite{gundersen_reproducible_2018,haibe-kains_transparency_2020,li_trustworthy_2023}. Gundersen and Kjensmo, in one of the most detailed examinations of reproducibility in AI to date \cite{gundersen_state_2018}, systematically surveyed 400 papers for a range of open science practices. They found that only about a third of papers share test datasets, only 8\% share source code, and only a single paper shared training, validation and test sets along with results. We are not aware of more recent systematic surveys of this kind (nor do we attempt this here), but the increasing trend of corporate releases with glossy blog posts replacing peer-reviewed scientific documentation provides little reason for optimism.

Openness is perhaps especially important for today's breed of instruction-following text generators, of which ChatGPT is the best known example. The persuasiveness of these language models is due in large part to an additional reinforcement learning component in which text generator output is pruned according to a reward function that is based on human feedback \cite{ziegler_fine-tuning_2020,ouyang_training_2022,cohen_dynamic_2022}, using insights from early work on evaluative reinforcement \cite{knox_tamer_2008, warnell_deep_2018,lambert_illustrating_2022}. Human users appear to be highly susceptible to the combination of interactivity and fluid text generation offered by this technology. The ubiquity of ChatGPT interfaces makes it easy for anyone today to try out some prompt engineering (while freely providing further training data to OpenAI) — but it does not allow one to gain a critical and holistic understanding of the constraints and capabilities of such systems, nor of their risks and harms. For true progress in this domain, we will need open alternatives.

In this paper, we survey alternatives to ChatGPT and assess them in terms of openness of data, models, documentation and access methods. The aim of our survey is threefold: to sketch some of the major dimensions along which it is useful to assess openness and transparency of large language models; to provide a view of the state of the art in open source instruction-tuned text generation; and to contribute towards a platform for tracking openness, transparency and accountability in this domain. 

\subsection{Previous work}
Existing work reviewing and comparing large language models falls into two categories: informal lists and structured surveys. Informal lists are crowd-sourced pointers to available resources, from open RLHF datasets\footnote{\href{https://github.com/yaodongC/awesome-instruction-dataset}{github.com/yaodongC/awesome-instruction-dataset} } to open examples of instruction-tuned text generators.\footnote{\href{https://github.com/nichtdax/awesome-totally-open-chatgpt/blob/main/README.md}{github.com/nichtdax/awesome-totally-open-chatgpt} } Systematic surveys of instruction-tuned language models are still rare and mostly focus on comparing model capabilities and performance, e.g., of ``augmented language models'' \cite{mialon_augmented_2023} and language models for writing code \cite{xu_systematic_2022} (not our focus here). Complementary to our focus on degrees of openness in instruction-tuned models, a recent survey of generative AI systems more broadly focuses on gradience in release methods, from closed to staged to fully open \cite{solaiman_gradient_2023}.

An important development in this domain the introduction of data statements \cite{mcmillan-major_data_2023} and model cards \cite{mitchell_model_2019}. These are structured documents that help creators document the process of curating, distributing and maintaining a dataset or model, and that help users to critically judge underlying assumptions, potential risks and harms, and potential for broader use. These resources have seen considerable uptake in the scientific community, though their adoption by for-profit entities lags behind. 

The risks of relying on proprietary solutions has spurred the development of several more open alternatives. For instance, the Bloom collaboration \cite{bigscience_workshop_bloom_2023} is a team science project of unprecedented magnitude. It has trained and open-sourced a large language model based on a collection of almost 500 HuggingFace datasets amounting to 1.6TB of text and code in 46 spoken languages and 13 programming languages. \cite{laurencon_bigscience_2022,muennighoff_crosslingual_2022}. A related initiative is The Pile \cite{gao_pile_2020}, a 800GB dataset of English text that serves as pre-training data for language models by EleutherAI \cite{phang_eleutherai_2022}. Meta AI's LLaMA \cite{touvron_llama_2023} provides researchers with access to a series of base models trained on data claimed to be `publicly available'. It should be noted that none of these initiatives have undergone rigorous peer-review or data auditing at this point, and that claims of openness do not cancel out problems, legal or otherwise.

In recent years, the private company HuggingFace has emerged as an important hub in the open source community, bringing together developers and users of projects in machine learning and natural language processing. It offers infrastructure for hosting code, data, model cards, and demos \cite{mcmillan-major_reusable_2021}. It also provides a widely used setup for automated evaluation, generating leaderboards and allowing quick comparison on a number of automated metrics, making it somewhat of a balancing act between offering incentives for documentation and for SOTA-chasing \cite{church_emerging_2022}. Our focus here is not performance evaluation of the kind offered by leaderboards; instead it is to survey degrees of openness in the fast-evolving landscape of text generators.

\section{Method}

We survey open-source instruction-tuned text generators and evaluate them with regard to openness, scientific documentation, and access methods. Since any survey in this fast-growing field deals with moving targets, we focus here mainly on dimensions of enduring relevance for transparency and accountability. An up to date list of all models surveyed can be found at \href{https://osf.io/d6fsr}{osf.io/d6fsr}.

\begin{table}[b]
\vspace{-10pt}
\begin{tabular}{|l|l|l|}
\hline
Availability & Documentation & Access methods \\ \hline
Open code    & License       & Package        \\
LLM data     & Code          & API            \\
LLM weights  & Architecture  &                \\
RLHF data    & Preprint      &                \\
RLHF weights & Paper         &                \\
             & Data sheet    &               \\ \hline
\end{tabular}
\caption{\textmd{Overview of the 13 assessment features.}}
\end{table}

\subsection{Requirements}

The target breed of models in focus here is characterized by the following two features: its architecture is at base a large language model with reinforcement learning from human feedback (LLM + RLHF) and it aims for openness and transparency (along degrees we quantify). Projects are not included if they are as  proprietary and undocumented as ChatGPT (like Google's Bard), or if they merely provide a front-end that calls some version of ChatGPT through an OpenAI API (like Microsoft's Bing). We explicitly include small-scale projects and projects that are in early stage development if they are open, sufficiently documented, and released under an open source license. Querying academic search engines and open code repositories, we find at least 15 projects that have sprung up in the last six months alone. 

\subsection{Survey elements}
We assess projects on 13 features divided over three areas (Table 1): availability, documentation, and access methods. For each feature, we document openness along a scale from maximum to partial to no openness and transparency. For licenses, only systems that are fully covered by a true open-source licence count as maximally open, less permissive or partial licensing counts as partially open, and non-open or unclear licensing situations count as closed. Figure 1 shows a snapshot of 15 projects assessed for all features, with degrees of openness colour-coded (\checkmark, $\sim$ , $\times$). Please refer to the data repository for more information about how each feature is evaluated, and for a more up to date listing.

\section{Results}
Projects roughly fall into two categories. First, small, relatively bare bones projects that only provide source code and build on existing large language models. These projects often cannot share information on architecture, training data, and documentation because they inherit closed-source data from the LLMs they build on. They usually also do not provide APIs or other user interfaces. However, some of such small projects do come with high-quality documentation and some build only on explicitly open LLMs. What such small projects lack in performance, they make up in utility for the open source community as they can provide useful entry points to learning about LLM+RLHF tools.

We also identify a handful of projects backed by larger organisations, which aim to offer similar features to proprietary tools such as ChatGPT but are open-sourced and well documented. Two such initiatives top our list of open-source alternatives to ChatGPT: bigscience-workshop's xmtf tool building on the BLOOMZ and mT0 models (sponsored by HuggingFace) and LAION-AI's OpenAssistant based on an open, crowd-sourced RLHF training dataset (oasst1). Open Assistant also features a text-based and graphical user interface as well as a web resources for crowd-sourcing training data. We also found that several projects are not as open as they initially seemed to be, with many of them merely wrappers of closed models.

We observe three recurring issues in the area of availability and documentation. \textit{Inheritance of undocumented data.} Many tools build on existing large language models (which we here call base models) and inherit the undocumented datasets (often web-scraped and often of dubious legality) these base models are trained on. 

\textit{Training data of RLHF component is not shared.} Building RLHF training datasets requires labour-intensive work by human annotators. The lack of RLHF training data is a major performance bottleneck for smaller research teams and organisations, and hampers reproducible research into the use of instruction-tuned text generators for conversational user interfaces. 

\textit{Papers are rare, peer-review even rarer.} Most projects reviewed here follow the corporate `release by blog post' model. While there are some preprints, none of the systems we review is currently documented in a peer-reviewed paper. Habitually bypassing this important (albeit sometimes flawed) quality assurance mechanism allows systems to escape critical scrutiny and risks undermining scientific and ethical standards.

Some other patterns are worth noting. One is the rise of synthetic data especially for the instruction component. Prominent examples are Self-Instruct (derived from GPT3) \cite{wang_self-instruct_2023}, and Baize, a corpus generated by having ChatGPT engage in interaction with itself, seeded by human-generated questions scraped from online knowledge bases \cite{xu_baize_2023}. This stretches the definition of LLM + RLHF architectures because the reinforcement learning is no longer directly from human feedback but has a synthetic component, in effect parasitizing on the human labour encoded in source models. The consequences of using synthetic reinforcement learning data at scale are unknown and in need of close scrutiny. 

The derivative nature of synthetic datasets is probably one reason they are released specifically ``for research purposes only'' \cite{xu_baize_2023}, with commercial use strictly prohibited. This leads to an important wrinkle. Baize models and data are incorporated in several popular instruction-tuned text generators, including the Falcon family of models which bills itself as ready for ``research and commercial utilization''\footnote{Technology Innovation Institute, https://falconllm.tii.ae/, June 7, 2023} in direct violation of Baize's prohibition against commercial use. This is merely one example of the complex dependencies embedded in these tools, and the legal quagmires obscured by simple claims of `openness'.

\section{Discussion}
The goal of this short paper has been to provide a critical review of degrees of openness in the fast-moving field of instruction-tuned large language models. We have found projects at varying stages of implementation, documentation, and useability. Most of them offer access to source code and some aspects of pre-training data, sometimes in legally ambiguous ways. Data from the reinforcement learning step, crucial to the simulation of instruction-following in these interfaces, is more elusive, provided by at best half of the initiatives. Strikingly, only a handful of projects are underpinned by a scientific write-up and none of them have as yet undergone scientific peer review. 

There are many shades of openness \cite{solaiman_gradient_2023}, yet all of the projects surveyed here are significantly more open than ChatGPT. ChatGPT was announced in a company blog post and rolled out to the public with an interface designed to capture as much free human labour as possible, but without any technical documentation. (The RLHF component, arguably the biggest differentiator for the instruction-following behavior, was sketched in \cite{ouyang_training_2022}, though without data.) Its follow-up GPT-4 continues OpenAI's tradition of openness in name only: it comes with an evaluation framework that primarily benefits the company yet contains the absolute minimum of technical documentation. In particular, an unreviewed preprint distributed by OpenAI and billed as a ``technical report" \cite{openai_gpt-4_2023} mostly provides cherry-picked examples and spends more space on crediting company workers for blog post content, communications, revenue, and legal advice than on actual technical details. (Companies like OpenAI sometimes give ``AI safety" as a pretext for closedness; this is hard to take seriously when their own public-facing proprietary models provide clear and present harms  \cite{gebru_statement_2023}.)

How can we foster more openness and accountability? First, incentives need changing. In high-stakes AI research, data work is often seen as low-level grunt work \cite{sambasivan_everyone_2021} and incentive structures generally encourage a `move fast and break things' mentality over careful scientific work \cite{rogers_changing_2021}. But work that documents data provenance and traces harmful impacts  \cite{birhane_algorithmic_2021,schaul_inside_2023} deserves major scholarly and societal credit. Here, AI and NLP might benefit from work in software engineering and infrastructure, where strong frameworks already exist to foster accountability for datasets \cite{liang_advances_2022,paullada_data_2021,hutchinson_towards_2021}. Interactive model cards \cite{crisan_interactive_2022} offer a promising step towards a human-centered approach to documentation.

Second, corporate capture and user lock-in are well-known strategies by which companies exercise control over scientific results and research infrastructure. In the age of large language models, this is amplified by the possibility to extract human labour and repackage it in amiable conversational formats. Openness not only aligns with principles of sound and ethical scholarship \cite{spirling_why_2023}; it also safeguards transparent and reproducible research \cite{nahar_collaboration_2022,muller_forgetting_2022}. Recent work on legal datasets offers an example in responsible data curation with insights that may be more broadly applicable \cite{henderson_pile_2022}.

Third, technology is never a \textit{fait accompli} unless we make it so. It is one of the achievements of publicly funded science that it can afford to not jump on the bandwagon and instead make room for reflection \cite{bender_dangers_2021,birhane_towards_2021}. Today's language technology landscape offers ample opportunities for what philosopher Ivan Illich has called \textit{counterfoil research}: ``Counterfoil research must clarify and dramatize the relationship of people to their tools. It ought to hold constantly before the public the resources that are available and the consequences of their use in various ways. It should impress on people the existence of any trend that threatens one of the major balances of which life depends'' \cite{illich_tools_1973}. Among the consequences of unleashing proprietary LLM + RLHF models are untold harms to workers exploited in labeling data; energy demands of computational resources \cite{luccioni_estimating_2022}; and tidal waves of plausible-looking text generated without regard for truth value (technically, bullshit \cite{frankfurt_bullshit_2009}). 

One possible outcome of the kind of deeper understanding fostered by openness is a call for responsibly limited technology \cite{illich_tools_1973,mcquillan_resisting_2022}. The spectre of regulation (a key way to keep corporate powers in check) is a powerful incentive for companies to keep things proprietary and so shield them from scrutiny. The systems we have surveyed here provide elements of a solution. Open to various degrees, they provide ways to build reproducible workflows, chart resource costs, and lessen reliance on corporate whims.

\section{Conclusion}

Openness is not the full solution to the scientific and ethical challenges of conversational text generators. Open data will not mitigate the harmful consequences of thoughtless deployment of large language models, nor the questionable copyright implications of scraping all publicly available data from the internet. However, openness does make original research possible, including efforts to build reproducible workflows and understand the fundamentals of LLM + RLHF architectures. Openness also enables checks and balances, fostering a culture of accountability for data and its curation, and for models and their deployment. We hope that our work provides a small step in this direction. 

\begin{acks}
This research is funded by Dutch Research Council (NWO) grant 016.vidi.185.205 to MD. For the purpose of Open Access the authors have applied a CC BY public copyright licence to any Author Accepted Manuscript version arising from this submission.
\end{acks}

\bibliographystyle{ACM-Reference-Format}
\bibliography{elpaco_shared}

\end{document}